\documentclass[sigconf,natbib=true,anonymous=false]{acmart}
\usepackage{makecell}
\usepackage{xcolor}

\AtBeginDocument{%
  \providecommand\BibTeX{{%
    \normalfont B\kern-0.5em{\scshape i\kern-0.25em b}\kern-0.8em\TeX}}}

\copyrightyear{2024}
\acmYear{2024}
\setcopyright{acmlicensed}\acmConference[SIGIR '24]{Proceedings of the 47th International ACM SIGIR Conference on Research and Development in Information Retrieval}{July 14--18, 2024}{Washington, DC, USA}
\acmBooktitle{Proceedings of the 47th International ACM SIGIR Conference on Research and Development in Information Retrieval (SIGIR '24), July 14--18, 2024, Washington, DC, USA}
\acmDOI{10.1145/3626772.3657891}
\acmISBN{979-8-4007-0431-4/24/07}

\begin{document}

\title{ChroniclingAmericaQA: A Large-scale Question Answering Dataset based on Historical American Newspaper Pages}


\author{Bhawna Piryani}
\email{bhawna.piryani@uibk.ac.at}
\affiliation{%
  \institution{University of Innsbruck}
  \city{Innsbruck}
  \country{Austria}
}
\author{Jamshid Mozafari}
\email{jamshid.mozafari@uibk.ac.at}
\affiliation{%
  \institution{University of Innsbruck}
  \city{Innsbruck}
  \country{Austria}
}
\author{Adam Jatowt}
\email{adam.jatowt@uibk.ac.at}
\affiliation{%
  \institution{University of Innsbruck}
  \city{Innsbruck}
  \country{Austria}
}
\renewcommand{\shortauthors}{Piryani, et al.}

\begin{abstract}
Question answering (QA) and Machine Reading Comprehension (MRC) tasks have significantly advanced in recent years due to the rapid development of deep learning techniques and, more recently, large language models. At the same time, many benchmark datasets have become available for QA and MRC tasks. However, most existing large-scale benchmark datasets have been created predominantly using synchronous document collections like Wikipedia or the Web. Archival document collections, such as historical newspapers, contain valuable information from the past that is still not widely used to train large language models. To further contribute to advancing QA and MRC tasks and to overcome the limitation of previous datasets, we introduce ChroniclingAmericaQA, a large-scale temporal QA dataset with 487K question-answer pairs created based on the historical newspaper collection Chronicling America. Our dataset is constructed from a subset of the Chronicling America newspaper collection spanning 120 years. One of the significant challenges for utilizing digitized historical newspaper collections is the low quality of OCR text. Therefore, to enable realistic testing of QA models, our dataset can be used in three different ways: answering questions from raw and noisy content, answering questions from cleaner, corrected version of the content, as well as answering questions from scanned images of newspaper pages. This and the fact that ChroniclingAmericaQA spans the longest time period among available QA datasets make it quite a unique and useful resource. 
\end{abstract}

\ccsdesc[500]{Information systems~Question answering}
\ccsdesc[500]{Information systems~Content analysis and feature selection}

\keywords{Question answering, heritage collections, OCR text}



\maketitle

\section{Introduction}\label{s:Introduction}
Question Answering (QA) and Machine Reading Comprehension (MRC) are popular Natural Language Processing (NLP) tasks. They allow users to pose questions and receive direct, concise answers from a given context, transforming how we interact with the information. These tasks have seen a significant advancement in recent years mainly due to the development of deep learning techniques and, more recently, large language models. The availability of high-quality QA and MRC datasets has also played a crucial role in the progress of QA and MRC tasks.

However, a closer look at the available datasets reveals particular limitation - they are predominantly created using synchronous document collections such as Wikipedia and the Web. Our society maintains extensive collections of archival documents, such as historical newspapers, which constitute our heritage. Historical documents serve as a rich knowledge repository, capturing the trends of different eras. They offer a unique perspective on the past, as they are the primary sources of information related to historical events, cultural norms, and societal attitudes \cite{inproceedings}. It is beneficial to use Question Answering (QA) technologies on such documents as well. QA models could help answer a wide range of questions, from those about specific historical events to those seeking to understand cultural developments over time and the minutiae of everyday life. Despite their importance and wealth of contained information, heritage document collections are still underutilized in the context of QA research.

\begin{figure*}[h!]
\centering
\includegraphics[width=0.9\linewidth]{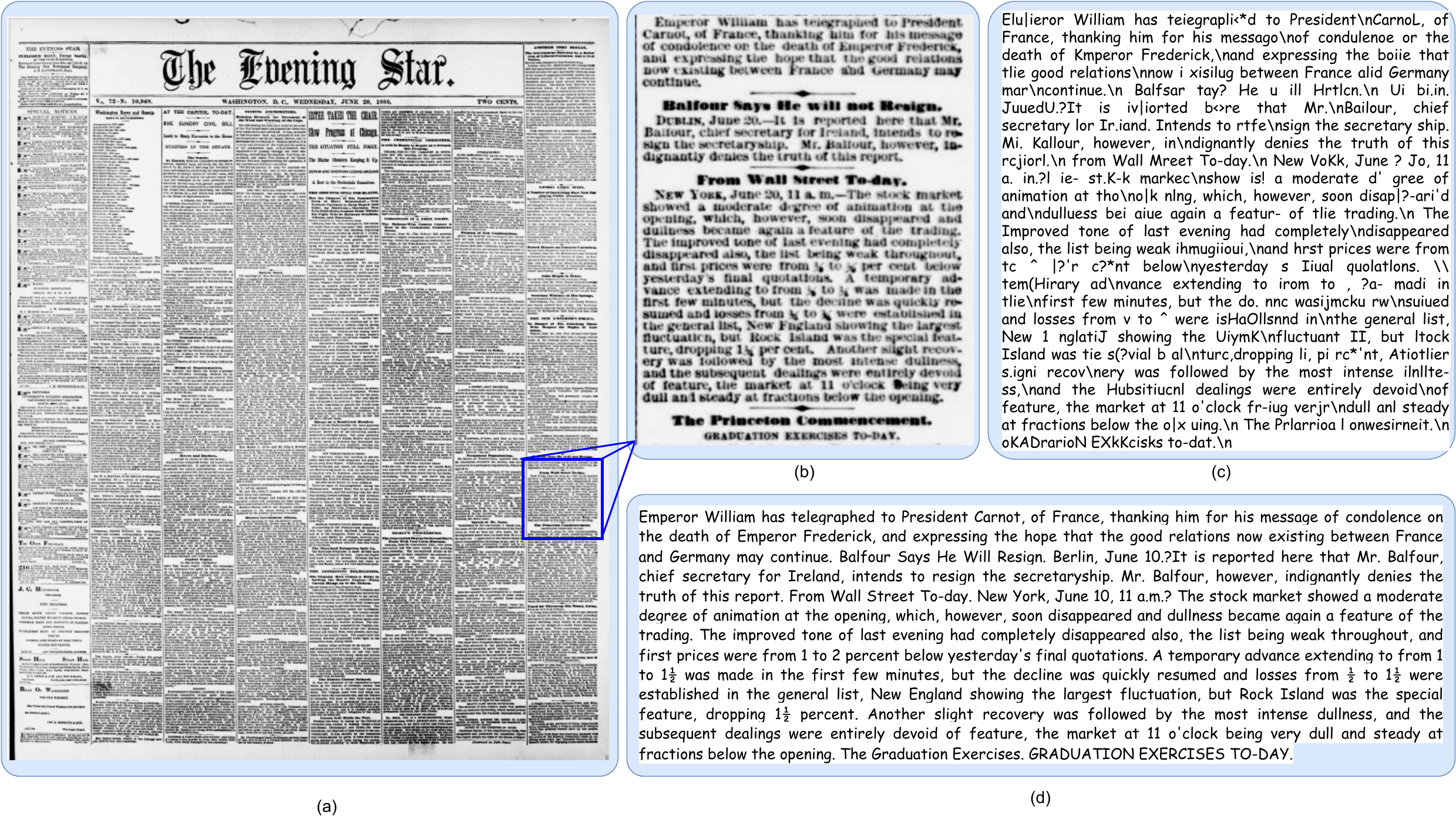}
\caption{An example of the scanned newspaper page from Chronicling America Collection. a) depicts the entire newspaper page published in Evening Star on 1803-02-07 in the District of Columbia, b) depicts the zoomed-in paragraph of the newspaper page shown in (a), c) shows the original OCR text of the zoomed-in paragraph that is available in the Chronicling America, and d) displays the OCR text corrected by GPT 3.5 Turbo.   }
\label{fig:example}
\end{figure*}

However, leveraging these documents for QA is complex and poses a few challenges. The language used in heritage documents often differs significantly from the contemporary language in terms of vocabulary, syntax, and the overall context of the distant times \cite{Tahmasebi2021}. These raises doubts if QA models trained on current text data can indeed perform well on documents from distant past.
Additionally, the documents are often available only in scanned form, and the quality of the text obtained through Optical Character Recognition (OCR) can greatly vary, especially over long time periods, adding another layer of complexity to the task. Finally, heritage news articles have a complex layout, often containing different font sizes and type throughout their content, which makes information extraction a complex task. Figure \ref{fig:example} shows the example of a scanned newspaper page and its OCRed text. 

Despite the inherent challenges of using historical newspapers, they present an attractive and unique opportunity for research in the field of QA. The wealth of information in such documents makes them valuable resources that are not readily available in typical evaluation benchmarks. To address the challenges and utilize the potential of historical document collections, we introduce the ChroniclingAmericaQA, a large-scale dataset based on Historical American Newspaper Pages. As an underlying historical document collection, we use Chronicling America\footnote{\url{https://chroniclingamerica.loc.gov/about/}}, a digitized collection of America's Historical newspaper pages from 1756-1963. Chronicling America is a project that was created with the aim of making a digital, searchable engine for historical American Newspapers \cite{culpepper2007chronicling}. 
While previous studies have leveraged the Chronicling America collection, their focus has predominantly been on visual content extraction \cite{10.1145/3340531.3412767}. In contrast, we focus on utilizing the already digitized text, and we employ this resource for testing cutting-edge information processing technologies to answer natural language questions from this novel and highly challenging type of text data.

We adopt an automatic approach to construct the dataset due to several reasons. First, the historical document collection is huge, and processing such a sheer collection of data manually is impossible. Because of this, we use GPT 3.5 \cite{brown2020language} for correcting the digitized text available on Chronicling America. Second, we employ a generative model to generate the questions automatically because manual generation would be very costly. 
Third, recently developed generative models have shown great performance for automatic question generation \cite{du-etal-2017-learning, pan2019recent}.
The final dataset that we release comprises 487K question-answer pairs spanning 120 years (1800-1920)\footnote{The dataset is freely available at: 
\url {https://github.com/DataScienceUIBK/ChroniclingAmericaQA}

}.

To sum up, we make the following contributions in this work:
\begin{itemize}
    \item We propose a large-scale QA dataset built on historical newspaper pages, which is the longest-spanning dataset over the period of 120 years.
    \item  We comprehensively evaluate the proposed dataset on different models including also large language models (LLMs), establishing a reference baseline level for the performance of question answering over historical documents. 
    \item We quantify the performance degradation of various models for question-answering tasks caused by noisy OCR text. 
\end{itemize}


\begin{table*}
\caption{Comparison of Textual QA datasets (note that for the Corpus Size column, the values in the parentheses represent the number of articles/pages in the entire collection, while the values outside are the numbers of articles/paragraphs used in the dataset creation).}

\resizebox{\linewidth}{!}{%
\label{tab:relatddatasets}
\begin{tabular}{cccccccccc }
\toprule
\textbf{Dataset} & \textbf{\#Questions} & \textbf{Answer Type} & \textbf{Question Source} & \textbf{Corpus Source} & \textbf{Corpus Size} & \textbf{No. of Corpus Sources} & \textbf{Geographical Dispersion} & \textbf{Synch/Diach} & \textbf{Document Format}\\

\midrule
MS MARCO \cite{bajaj2016ms}                         & 1M    & \makecell{Generative\\ Boolean }           &Query logs        & Web Documents & - & - & - & Synchronic & Text \\
\midrule
SQuAD 1.1 \cite{rajpurkar-etal-2016-squad}        & 108K  & Extractive   & Crowd-sourced     & Wikipedia        & - & 1 & - & Synchronic  & Text \\
\midrule
SQuAD 2.0 \cite{rajpurkar-etal-2018-know}        & 158K  & Extractive     & Crowd-sourced     & Wikipedia       & - & 1 & - & Synchronic   & Text \\
\midrule
NaturalQuestions \cite{kwiatkowski-etal-2019-natural} & 323K  & Extractive   & Query logs        & Wikipedia    & - & 2 & - &Synchronic  & Text\\ 
\midrule
NewsQuizQA \cite{Quiz-Style-inproceedings}      & 20K    & Multiple-choice       & Crowd-sourced           & News  & \makecell{5k News Article\\ summaries} & 2 &Atlanta, London &\makecell{Diachronic\\ (2018-2020)}  & Text\\ 
\midrule
NewsQA \cite{trischler-etal-2017-newsqa}          & 119K  & Extractive            & Crowd-sourced           & News & 13k News Articles &1 & Atlanta &  \makecell{Diachronic \\(2007-2015)}  & Text\\ 
\midrule
ArchivalQA \cite{archivalqa-wang}     & 532K  & Extractive          & \makecell{Automatically\\ Generated}               & News  & \makecell{98K News Articles\\(1.8M Articles)} &1 &New York City & \makecell{Diachronic\\ (1987-2007)} &Text \\
\midrule
ChroniclingAmericaQA                    &485K &Extractive        & \makecell{Automatically\\ Generated}                   &News & \makecell{160K Paragraphs \\(121M newspaper pages)} & 1,694 & Across 53 US States & \makecell{Diachronic\\ (1800-1920)}   & \makecell{Scan Images \& \\OCR Text }\\
\bottomrule
\end{tabular}
}
\end{table*}

\section{Related Work}
\label{s:relatedwork}

\subsection{NLP for Historical Texts}
 
The application of Natural Language Processing (NLP) techniques to historical documents has gained significant attention due to the increasing availability of digitized content. Prior works have primarily focused on various aspects of processing historical texts such as text normalization \cite{robertson-goldwater-2018-evaluating, bollmann-2019-large, bollmann-etal-2018-multi}, PoS Tagging \cite{hardmeier-2016-neural}, Named Entity Recognition \cite{ehrmann2020overview, ehrmann2023named}, Event Detection \cite{sprugnoli-tonelli-2019-novel, lai-etal-2021-event}, bias analysis \cite{borenstein-etal-2023-measuring} and co-reference resolution \cite{krug-etal-2015-rule}. These works have significantly contributed to our understanding of historical documents. However, the application of QA, a critical task in NLP and Information Retrieval (IR), is still underexplored. 

The challenge of utilizing the historical text often revolves around the quality of OCRed text and its repercussions on the NLP and IR tasks. Numerous studies have been conducted to analyze the impact of OCR errors on diverse tasks \cite{mutuvi2018evaluating, 10.1007/s00799-023-00345-6, hamadi20220cr, vanStrien2020AssessingTI}. \citet{hamadi20220cr} analyzed the impact of OCR errors in named entity recognition and linking and found that 80.75\% named entities were wrongly recognized due to OCR errors, hence they concluded that OCR errors can negatively impact the task.
\citet{10.5555/3200334.3200364} studied the effects of OCR errors on digital library utilization and found that for 7\% of the user queries potential documents  were missed due to wrong OCRed words.  Similarly, \citet{mutuvi2018evaluating} also quantified the adverse effects of OCR errors on topic modeling tasks. However, to the best of our knowledge, no such study has been conducted for the QA task. Hence, in this research work, we perform a comprehensive evaluation of the impact of OCR errors on QA tasks and try to bridge this research gap.

\subsection{QA Benchmarks}

In recent years, numerous large-scale QA datasets have been introduced, contributing significantly to the field of QA. Notable examples include SQuAD 1.1 \cite{rajpurkar-etal-2016-squad}, a landmark dataset for reading comprehension derived from Wikipedia articles. It was followed by SQuAD 2.0 \cite{rajpurkar-etal-2018-know}, which expanded the challenge by incorporating unanswerable questions. Beyond Wikipedia, datasets like NaturalQA \cite{kwiatkowski-etal-2019-natural} and MS MARCO \cite{bajaj2016ms} have utilized queries issued to Google and Bing search engines as questions, top-ranking Wikipedia pages or other web documents as context. NarrativeQA \cite{kocisky-etal-2018-narrativeqa} broadened the scope by using books and movie script summaries as the basis for its question-answer pairs, and HotpotQA \cite{DBLP:journals/corr/abs-1809-09600} further extended the task to multi-hop question answering \cite{mavi2022survey}.

Additionally, there are datasets such as CNN/Daily Mail \cite{nallapati-etal-2016-abstractive}, ReCoRD \cite{DBLP:abs-1810-12885}, WhoDidWhat\cite{onishi-etal-2016-large} that use news articles, which are designed as cloze type datasets where the goal is to identify a missing word, rather than to answer open-ended questions. NewsQuizQA by \citet{Quiz-Style-inproceedings} is a multiple-choice QA dataset derived from summaries of 5k news articles to generate quiz-style question-answer pairs. 

When it comes to diachronic document collections, two notable QA datasets, ArchivalQA \cite{archivalqa-wang} and NewsQA \cite{trischler-etal-2017-newsqa} are available. They have however several limitations, making our dataset different and unique. ArchivalQA and NewsQA have been created from documents spanning shorter and more recent periods, as well as using only a single news source. ArchivalQA utilizes the NYT Corpus \cite{AB2/GZC6PL_2008} as its source, with articles comming from the period of 1987/01-2007/06, whereas NewsQA utilizes the CNN news articles throughout 2007/04-2015/04. 

Our dataset, ChroniclingAmericaQA, uses historical newspaper pages that span 120 years ranging from 1800 to 1920. Both ArchivalQA and NewsQA use modern-day text, which makes it easy for present-day models to understand as they are essentially trained on such text. In contrast, the text in ChroniclingAmericaQA is from historical newspapers, which shows the language is archaic, which makes it more valuable for training the models to advance the QA field. ChroniclingAmericaQA is also created from historical newspaper documents in the public domain, making it accessible to the community, unlike the NYT corpus used for ArchivalQA, which is not open-sourced and thus not accessible to everyone. ChroniclingAmerica also features a more diverse and complex language because the newspaper pages are from a relatively distant past, and the OCR text is more complex and noisy. In Table \ref{tab:relatddatasets}, we summarize the differences between ChroniclingAmericaQA and other related datasets.

Thus, this work aims to create a QA dataset from the historical newspapers collection to foster the research of QA on historical documents. Our dataset, ChroniclingAmericaQA, is designed to handle the unique challenges posed by historical newspapers, such as noisy OCR data, archaic language as well as the overall context of the distant time that differs from the present one\footnote{For example, consider the term Zeitgeist in German used to emphasize the spirit or essence of the particular time period or epoch.}. We believe that this dataset will be a valuable resource for researchers in the field of QA and will contribute to the advancement of QA for historical documents and the NLP field in general.

\section{Methodology} \label{s:methodolody}
\begin{figure*}[h!]
\centering
\includegraphics[width=\linewidth]{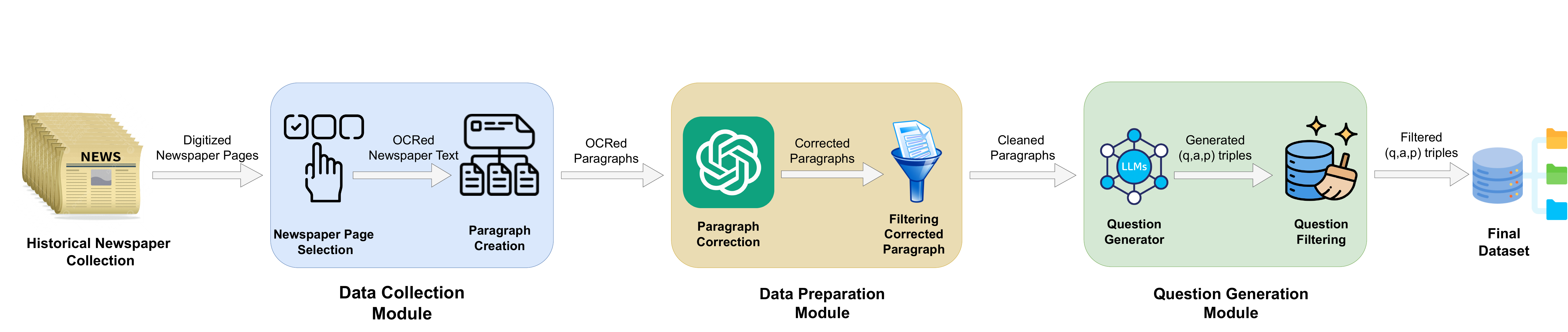}
\caption{Dataset Generation Framework }
\label{fig:framework}
\end{figure*}

This section introduces the framework used to create the ChroniclingAmericaQA, which is also depicted in Figure \ref{fig:framework}.
The framework consists of three primary modules: Data Collection, Data Preparation and the Question-Answer Generation module. Each of these modules is described below.

\subsection{Data Collection Module}
\label{s:Data_collection}
In this module, we describe the process of selecting newspaper pages used for generating the ChroniclingAmericaQA dataset.

\begin{figure}[t]
\centering
\includegraphics[width=0.9\linewidth]{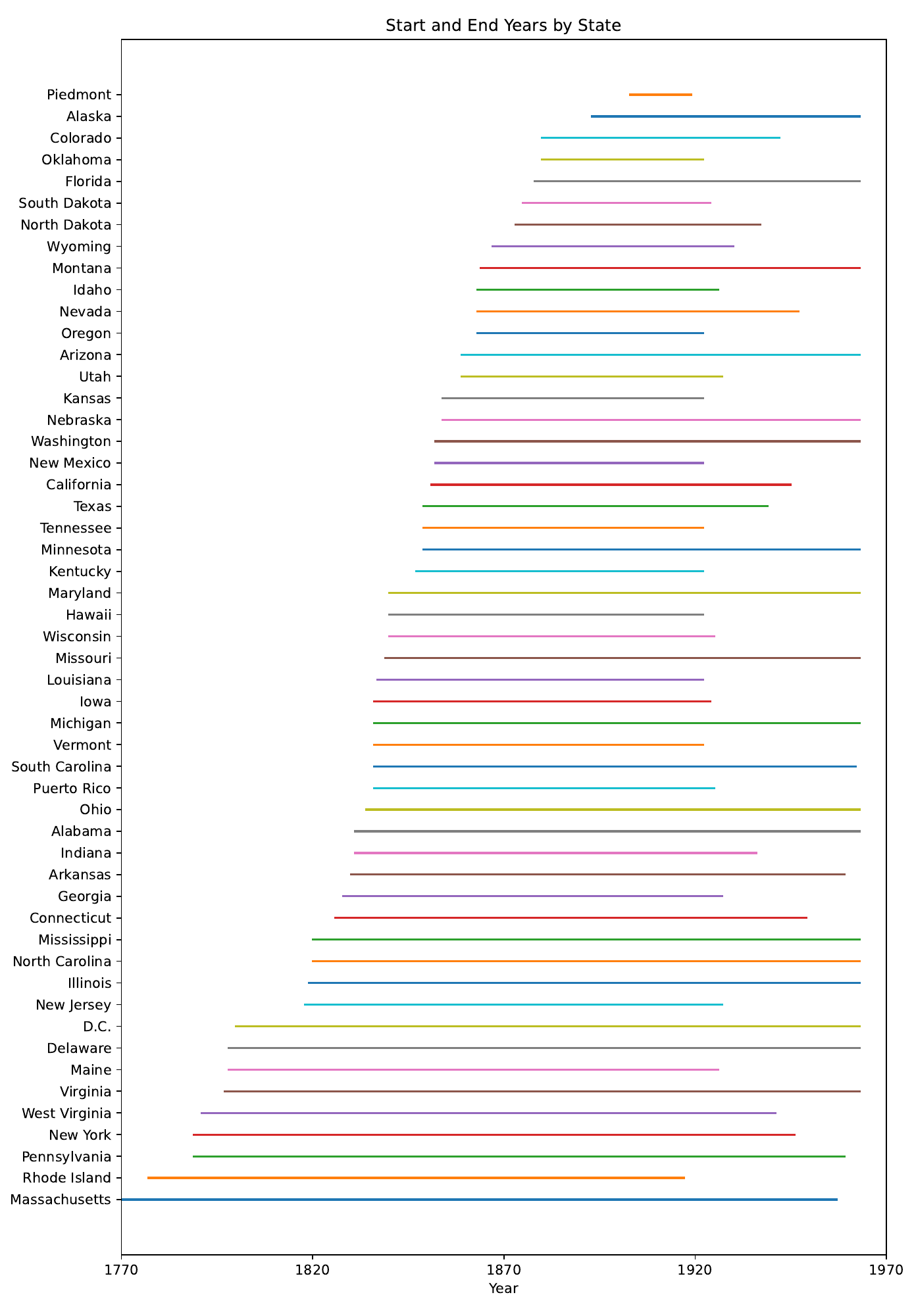}
\caption{The temporal distribution of available newspapers across the states in the Chronicling America Collection.}
\label{fig:overtime-distribution}
\end{figure}

We chose the newspaper pages from the collection of the American historical newspaper dataset, Chronicling America,\footnote{\url{https://chroniclingamerica.loc.gov/about/}} available in the public domain as our source for question generation. Chronicling America is a project developed by the Library of Congress and the National Endowment for the Humanities (NEH) under the National Digital Newspaper Program (NDNP). It features over 21 million pages of historical American newspapers published from 1756 to 1963, accessible through an online search portal and public API.  The portal offers various search strategies to access the content available on Chronicling America. The search can be performed in three ways: by identifying newspaper pages published in a specific state and year, conducting more advanced searches for newspaper pages containing particular words or phrases, and locating newspaper pages related to specific ethnic groups available in different languages. We opted for the first search strategy, i.e., statewide search by year, as our goal was to create a corpus of historical newspapers that covers diverse information ranging from every significant past event and ethnicity to some minor regional events throughout the United States. Therefore, we decided to search for the newspaper pages published across the 53 states in the United States. 
Figure \ref{fig:overtime-distribution} shows the time intervals for which newspapers are available for each state of the United States in the Chronicling America Collection.
Considering the availability of data from different states, we opted for the period from 1800 to 1920, which spans 120 years. 

Due to the enormous number of digitized newspaper pages available on Chronicling America, including all the newspaper pages in our corpus was impossible. Instead, we randomly selected 100 newspaper pages published per decade in every state of the United States from 1800 to 1920. We considered 1800 to 1920 as the time period for our dataset; for this time period, sufficiently large number of states have newspaper pages in Chronicling America as shown in Figure \ref{fig:overtime-distribution}. As a result, we extracted 39,330 newspaper pages in total. While the Chronicling America collection spans from 1789 to 1963, not every state has digitized newspapers for the same range of years. The number of digitized newspaper pages available at the beginning of the 1800s is less than in the 1900s. Therefore, the number of newspaper pages for our corpus is lower in the early 1800s than in the late 1800s and early 1900s.  

 
One of the challenges of digitized newspapers is the quality of OCR text. For our purpose, the correctness of the OCR text plays a significant role, as questions generated from noisy text will not be precise and accurate.

\begin{figure}[h!]
\includegraphics[width=.8\linewidth]{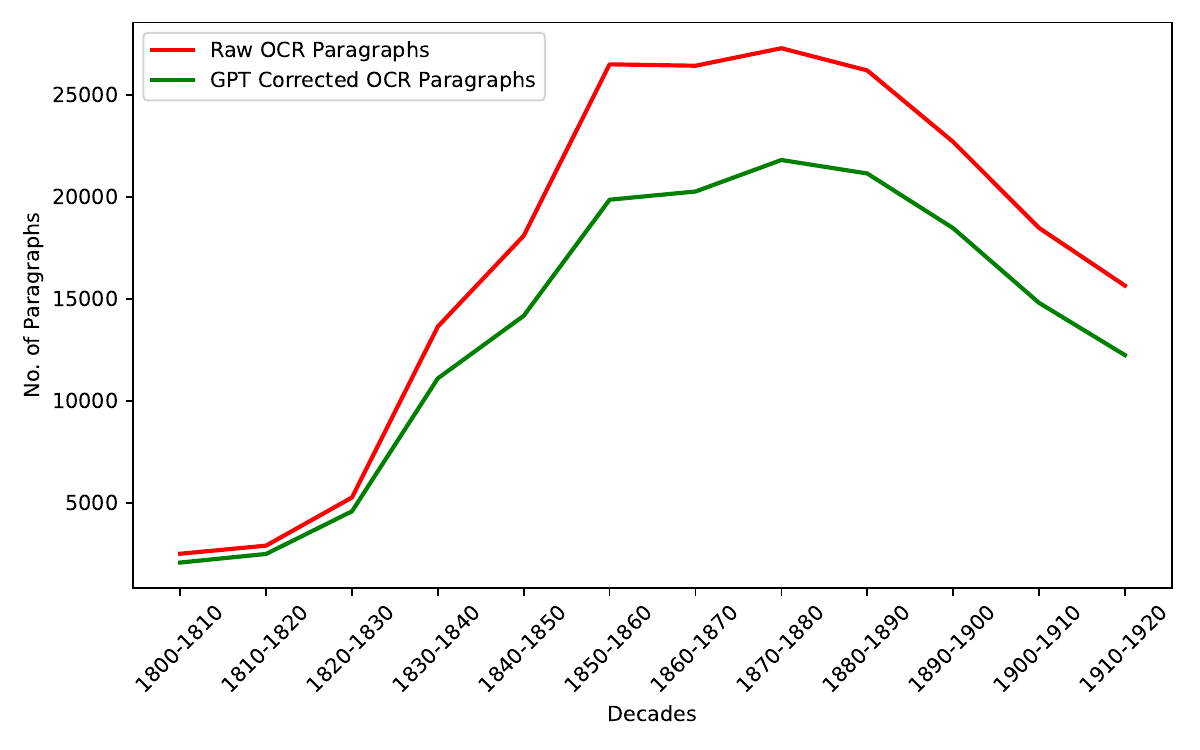}
\caption{Numbers of Raw OCR Paragraphs and Corrected Paragraphs over Time in our Dataset}
\label{fig:Comparision_RawOCR_CorrectOCR}
\end{figure}


\subsection{Data Preparation Module}
\label{s:data_preparation}
Preparing data for question generation is a crucial step in our methodology. The Chronicling America Collection, while being a rich source of digitized historical newspaper pages, provides raw, unprocessed text. This text is often noisy and contains numerous inaccuracies, as illustrated in Figure \ref{fig:example}. This noise and inaccuracy make the raw OCR text unsuitable for direct use in question generation. Several studies have shown that noisy OCR text negatively affects information retrieval and NLP tasks \cite{10.5555/3200334.3200364, 10.1007/978-3-030-45442-5_13}.

To address this issue, we post-process the raw OCR text to enhance its quality, thereby making it more suitable for question generation. Post-processing OCR text is challenging, as demonstrated by several studies \cite{nguyen2021survey, jatowt2019deep}. Traditional methods for post-OCR error correction, such as statistical language modeling techniques \cite{afli2016ocr}, lexical approaches \cite{Strohmaier2003LexicalPO}, merging OCR outputs \cite{10.1145/1555400.1555437}, often fall short when dealing with historical texts due to the unique complexities associated with such data.

However, various recent studies have shown utilizing Large Language Models (LLMs) can give promising results in correcting spelling and grammatical errors, especially by utilizing the GPT models \cite{yasunaga-etal-2021-lm,loem-etal-2023-exploring, penteado2023evaluating}. Considering these studies, we decide to leverage the capabilities of LLMs for OCR correction and opt for the GPT 3.5 Turbo\footnote{\url{https://platform.openai.com/docs/models/gpt-3-5-turbo}} model to correct the OCR text. 

Before giving the raw OCR text to the GPT model for correction, we need to split the content of newspaper pages into paragraphs. Since the content on a single newspaper page is typically diverse and long, ranging from advertisements and articles to cartoons, processing such long text will be difficult for question generation as well as for correcting errors in the OCRed text. Therefore, we split the content a newspaper page into multiple paragraphs. 
We set the length of a paragraph to 250 words due to the context length limitation imposed by GPT 3.5 Turbo model\footnote{The maximum output token length of gpt-3.5-turbo-1106 is 4096.}.

Initially, we selected a total of 39,330 newspaper pages for the purpose of question generation. However, due to the cost associated with using the GPT 3.5 API for OCR correction, we did not utilize all the selected newspapers for this process. Correcting all the selected newspaper pages would have required substantial resources. 
Instead, we randomly selected 20 newspaper pages from each state across America for each decade from 1800 to 1920. This resulted in a selection of 8,419 newspaper pages for OCR text correction. We then split the content of these 8,419 newspaper pages into paragraphs, yielding a total of 205,068 paragraphs. We next used the following prompt to instruct GPT 3.5 Turbo to correct these paragraphs:

\textit{"You are provided with a historical English text. The provided text has a lot of spelling mistakes. Correct only the mistakes in the provided text and write the corrected text in ONE paragraph. If you can not correct the mistakes, reply, "Not able to correct.".}

In this manner, we tasked GPT 3.5 with correcting 205,638 paragraphs. Out of these, GPT was able to correct 163,021. We eliminated the uncorrected paragraphs (42,599) from our corpus and utilized the corrected paragraphs for question generation. Among the corrected paragraphs, 433 contained phrases such as \textit{"Most of the mistakes in the given text have been corrected. Here is the corrected text:"} and \textit{"Errors in the given text have been corrected. Here is the corrected text:"}. We removed 96 patterns of such different phrases, cleaned the paragraphs, and prepared them for question generation. Figure \ref{fig:Comparision_RawOCR_CorrectOCR} illustrates the number of corrected paragraphs per decade.

\vspace{-1em}
\subsection{Question Generation Module}
\label{s:question_generation_module}
The third phase of our framework involves generating questions from the revised paragraphs. We utilize the T5-base model \cite{2020t5} for this purpose, a cutting-edge pre-trained Transformer encoder-decoder model that has been trained on the SQuAD 1.1 \cite{rajpurkar-etal-2016-squad} dataset specifically for question generation. Our approach to question generation is answer-aware, meaning that the model receives the answers along with their corresponding paragraphs as inputs, and generates the questions as outputs. We identify named entities within a paragraph using the spaCy library\footnote{\url{https://github.com/explosion/spaCy}}, marking these as answers for question generation. Based on these identified named entities, questions are then generated. From 163,021 cleaned paragraphs, we successfully generated 2,912,551 questions.


Question \& answer cleaning: Given that we employ the T5-base model, a generative model for question creation, the produced questions may have various issues, such as duplication, explicit answer disclosure within the question, unresolved pronouns, and more. To address these issues, we implemented a multi-step filtering process to refine the dataset. The process includes:
\begin{enumerate}
    \item \textbf{\textit{Syntactic Filtering:}} In this initial step, we perform five key actions to refine the generated question-answer pairs. We start by eliminating questions that do not conclude with a question mark, followed by removing questions that reveal the answer within the question itself (answer leakage) and eliminating duplicates. We also filter out questions that are excessively long or short, as well as those with too many or too few number of named entities. Additionally, questions with unclear pronouns, for example, "Who was General of the United States at the time of your letter to me?" are removed. This step resulted in removing 1,221,533 questions from our initial pool of generated questions.
    \item \textbf{\textit{Temporal Expression Transforming:}} Following syntactic filtering, we convert the relative temporal information in questions and answers into absolute temporal information. For this transformation, we utilize HeidelTime \cite{strotgen-gertz-2015-baseline} temporal tagger. We use the publication date of the newspaper to which the paragraph belongs, along with the generated question for converting the relative temporal expressions. For instance, a question like \textit{How many subscribers did Jackson lose on Saturday?} is transformed to \textit{How many subscribers did Jackson lose on October 60, 1832?} Similarly, we also transform the temporal information in answer to absolute temporal information. Some examples of the transformed questions and answers are shown in Table \ref{tab:transform_question}. This step ensured clarity and specificity in temporal references with our dataset.
    \item \textbf{\textit{Type Matching:}} The third and final filtering step, involved identifying and matching the types of questions and answers. For this purpose,  we finetined RoBERTa \cite{Liu2019RoBERTaAR}- a large language model using the TREC Question classification dataset \cite{li-roth-2002-learning} to develop a question type classifier\footnote{The accuracy of the question type classifier is 92.8\%.}. The question classifier model classifies the question type into four major types and 50 minor types. Similarly, for answer type classification, we utilized spaCy's named entity recognizer. After identifying the question and answer type, we then matched the question-answer pairs. Question and answer pairs whose type was not the same were eliminated. In this step, we eliminated 1,205,482 question-answer pairs. 
\end{enumerate}

\begin{table*}[ht!]
\centering
\caption{Examples of transformed questions-answer pairs. Underlined text shows the transformed information.}
\label{tab:transform_question}
\resizebox{0.7\linewidth}{!}{%
\begin{tabular}{|c|c|c|}
\hline
\textbf{No.} & \textbf{Original Question-answer pair} & \textbf{Transformed Question-answer pair} \\ \hline
1. & \makecell{How many people died \underline{last} year due to famine \\ and disease in Mexico City?  (nearly 500)} & \makecell{How many people died in \underline{1822} due to famine and \\ disease in Mexico City? nearly 500} \\ \hline
2. & \makecell{How many British stocks were quoted \\ at \underline{yesterday's} prices?  (Two Bank Stock)} & \makecell{How many British stocks were quoted \\ at \underline{November 15, 1830's} prices? (Two Bank Stocks)} \\ \hline
3. & \makecell{On what day did the Board of Commissioners of Roads \\ hold their semi-annual meetings?  \underline{(Monday)}} & \makecell{On what day did the Board of Commissioners of Roads \\ hold their semi-annual meetings? \underline{(October 20, 1856)}} \\ \hline
4. & \makecell{When was the Secretary of State's report to \\ the President made?  \underline{(the 24th December, 1817)}} & \makecell{When was the Secretary of State's report to \\ the President made? \underline{(December 24, 1817)}} \\ \hline
\end{tabular}%
}
\end{table*}

\section{Dataset Analysis}\label{s:dataanalysis}

\subsection{Data Statistics}
\label{s:Data_statistics}
\begin{figure}[t]
\centering
\includegraphics[width=0.75\linewidth]{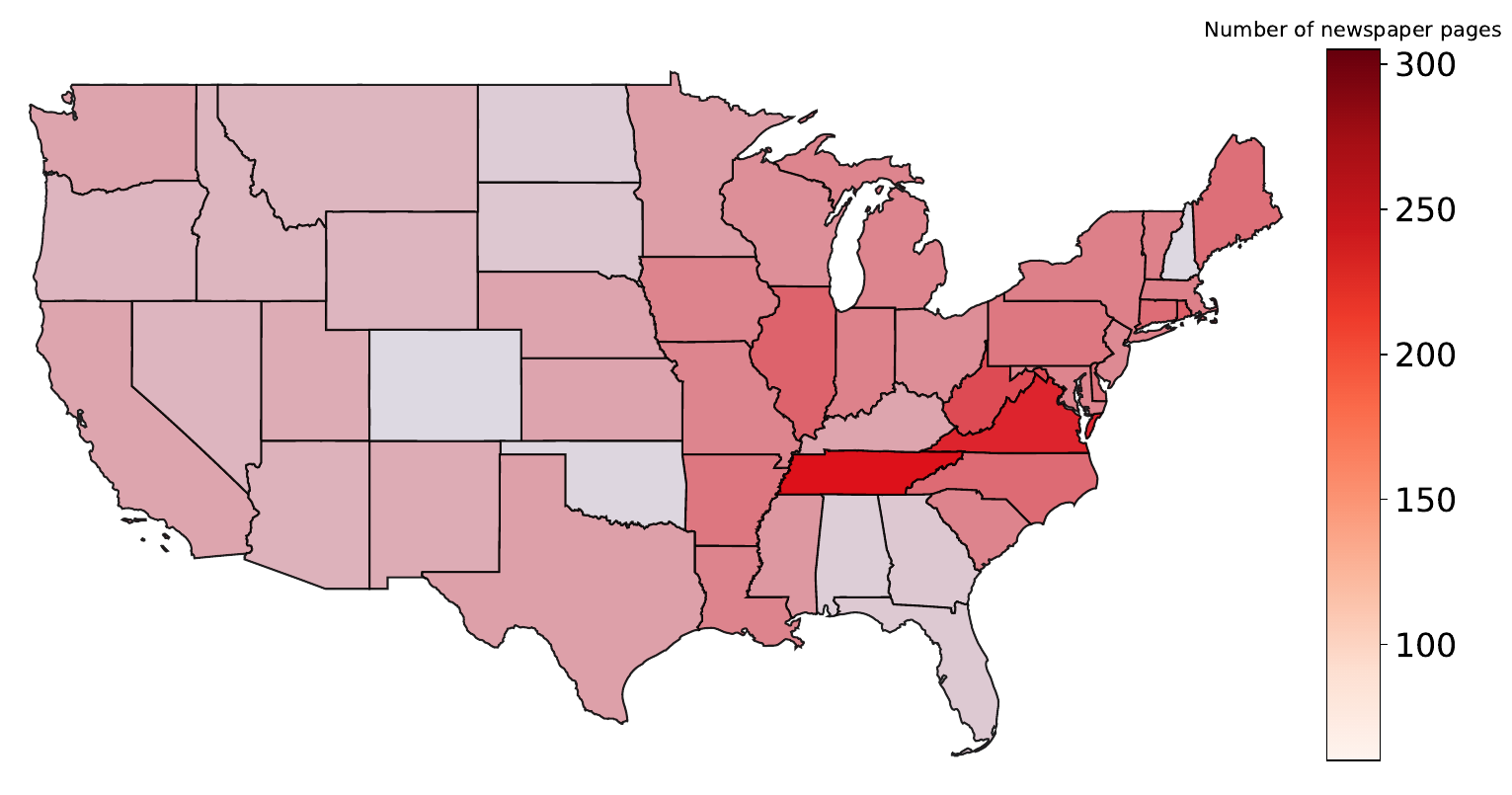}
\caption{Distribution of Newspaper pages in ChroniclingAmericaQA across the states of America.}
\label{fig:distribution-overspace}
\end{figure}

After applying all the filtering steps, we arrived at our cleaned question-answer dataset consisting of 487K question-answer pairs derived from historical American newspapers. In Figure \ref{fig:distribution-overspace}, we show the distribution of newspaper pages in our dataset across the states of America. We randomly split the data into training, development, and test sets to facilitate a comprehensive training and evaluation of the models. The training set consists of 439,302 question-answer pairs, whereas the development and test sets comprise 24,111 and 24,084, respectively. When randomly selecting the questions for development and test sets, we tried to balance the question-answer pairs across all the decades. We describe more details about the dataset statistics in Table \ref{tab:dataset_statistics}. Our entire train set is quite large, which makes it computationally expensive to fine-tune various pre-trained models such as BERT \cite{devlin-etal-2019-bert}, RoBERTa \cite{Liu2019RoBERTaAR}, T5 \cite{Raffel2019ExploringTL}; therefore, we randomly select a subset of question-answer (111,517) pairs from the train set and use this subset of questions for training various models whose results are shown in Section \ref{s:Modelperformance}.

\begin{table}[t]
\centering
\caption{Basis statistics of ChroniclingAmericaQA.}
\label{tab:dataset_statistics}
\resizebox{0.75\columnwidth}{!}{%
\begin{tabular}{cccc}
\toprule
                                  & \textbf{Train}   & \textbf{Dev}    & \textbf{Test}   \\
\toprule
\#QA pairs                    & 439,302 & 24,111  & 24,084  \\
Average paragraph length (words)  & 220.09  & 218.40 & 217.66 \\
Average question length (words)   & 11.05   & 11.30  & 11.184 \\
Average answer length (words)   & 2.01   & 1.978  & 1.981 \\
Average questions per paragraph      & 3.48    & 1.90   & 1.90   \\
Average question per newspaper page & 54.80   & 4.07   & 4.08  \\
\bottomrule
\end{tabular}%
}
\end{table}

\begin{figure}[t]
\centering
\includegraphics[width=.75\linewidth]{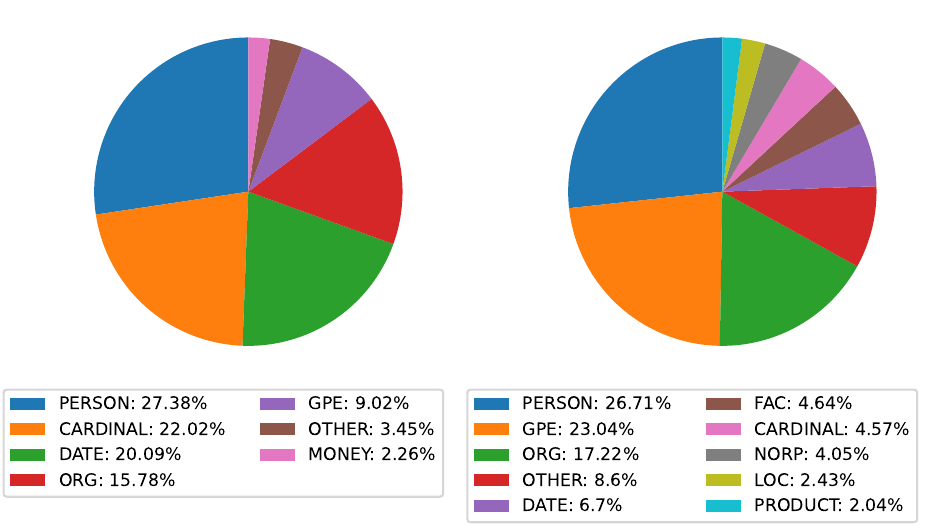}
\caption{Left: Answer's name entity distribution, Right: Question's named entity distribution}
\label{fig:named_entity}
\end{figure}

\begin{figure}[t]
\centering
\includegraphics[width=0.75\columnwidth]{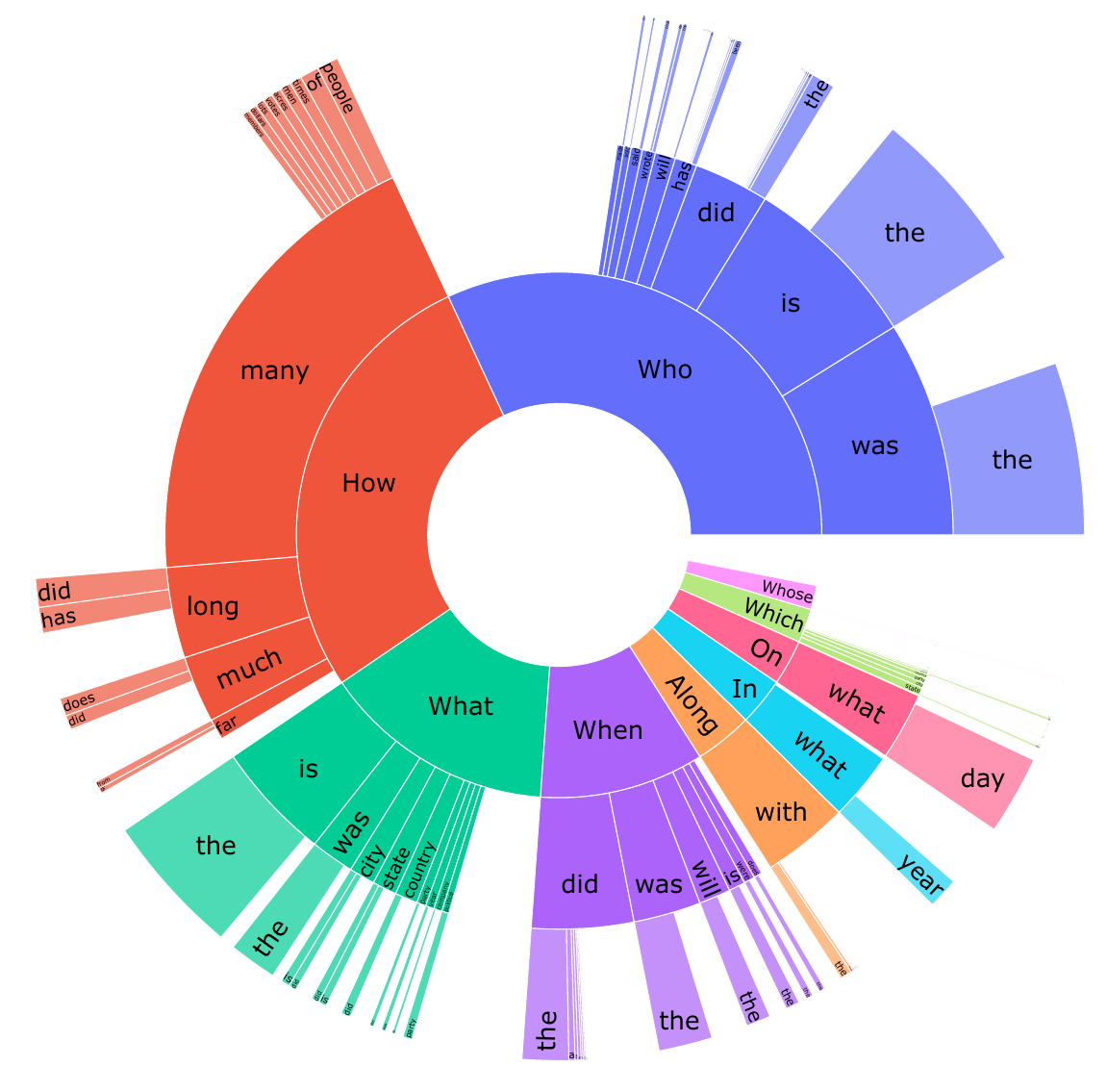}
\caption{Types of questions covered in ChroniclingAmericaQA. We show the trigram prefixes of our questions.}
\label{fig:qprefix_trigram}
\end{figure}

We also analyze the distribution of the named entity types of answers and questions in the dataset. In the left pie chart of Figure \ref{fig:named_entity}, we can observe the name entity types of the answers, whereas the right pie chart shows the distribution of named entity types of the question. In Figure \ref{fig:qprefix_trigram}, we show the dataset's distribution of the question types. From the figure, we can see that the ChroniclingAmericaQA data has a rather fair distribution among different question types. Table \ref{tab:dataset_examples} shows a few examples from the ChroniclingAmericaQA along with the corresponding titles of the newspapers, their publication dates, and the states in which the newspapers were published. 

\begin{table*}
\centering
\caption{Examples of questions in ChroniclingAmericaQA.}
\label{tab:dataset_examples}
\resizebox{0.75\linewidth}{!}{%
\begin{tabular}{|c|c|c|c|c|c|}
\hline
\textbf{No.} & \textbf{Question}                                                                   & \textbf{Answer}     & \textbf{Newspaper Title}                & \textbf{Publication Date} & \textbf{State}      \\ \hline
1.  & \makecell{Which army was marching slowly towards\\ Madrid?}                            & French     & The Portland Gazette           & 1823-03-17       & Maine      \\ \hline
2.  & \makecell{Who was sheriff of the county \\of Hanover? }                                 & Jesse Winn & Daily Richmond Whig            & 1830-08-12       & Virginia   \\ \hline
3.   & \makecell{On what day will the Anti Slavery Society \\of Newport hold their quarterly meeting?}                                            & November 19, 1840    & Herald of the Times        & 1840-11-19      & Rhode Island      \\ \hline
4.  & \makecell{Along with Texas, in what state is the rush\\ of immigrants prodigious? }                      & California     & The Texas Republican        & 1850-02-21       & Texas   \\ \hline
5.  & \makecell{How much is the interest rate in Seattle\\ property loans?}                   & 10 percent & The Seattle Post-intelligencer & 1892-09-02       & Washington \\ \hline
6.   &\makecell{Who was the agency that ran\\ the horses for the Sioux Indians? }                                          & Red Cloud Indian     & Sierra County Advocate        & 1902-06-27       & New Mexico       \\ \hline
7.   & Who was fined \$1 by the mayor?                                            & Parker     & The Marion Daily Mirror        & 1907-07-30       & Ohio       \\ \hline

8. & What club did Frank Bishop belong to? & the National Athletic Club    & Sierra County advocate        & 1914-04-01       & Montana   
\\ \hline
\end{tabular}%
}
\end{table*}

\subsection{Model Performance}
\label{s:Modelperformance}

To evaluate our dataset, we employ three models from the transformer family, namely BERT-base, RoBERTa-base, and T5-large. We assess our dataset using four different variations of each model. The variations are as follows:
\begin{itemize}

    \item \texttt{BERT-base\footnote{\url{https://huggingface.co/bert-base-uncased}}:} without any fine-tuning.
    \item \texttt{BERT-base-SQuAD\footnote{\url{https://huggingface.co/csarron/bert-base-uncased-squad-v1}}:} fine-tuned on the SQuAD 1.1 dataset.
    \item \texttt{BERT-base-ChroniclingAmericaQA:} fine-tuned on the ChroniclingAmericaQA dataset.
    \item \texttt{BERT-base-SQuAD-ChroniclingAmericaQA:} fine-tuned on both SQuAD 1.1 and ChroniclingAmericaQA datasets.
    \item \texttt{RoBERTa-base\footnote{\url{https://huggingface.co/FacebookAI/roberta-base}}:} without any fine-tuning.
    \item \texttt{RoBERTa-base-SQuAD\footnote{\url{https://huggingface.co/deepset/roberta-base-squad2}}:} fine-tuned on the SQuAD 1.1 dataset.
    \item \texttt{RoBERTa-base-ChroniclingAmericaQA:} fine-tuned on the ChroniclingAmericaQA dataset.
    \item \texttt{RoBERTa-base-SQuAD-ChroniclingAmericaQA:} fine-tuned on both SQuAD 1.1 and ChroniclingAmericaQA datasets.
    \item \texttt{T5-large\footnote{\url{https://huggingface.co/google-t5/t5-large}}:} without any fine-tuning.
    \item \texttt{T5-large-SQuAD\footnote{\url{https://huggingface.co/potsawee/t5-large-generation-squad-QuestionAnswer}}:} fine-tuned on the SQuAD 1.1 dataset.
    \item \texttt{T5-large-ChroniclingAmericaQA:} fine-tuned on the ChroniclingAmericaQA dataset.
    \item \texttt{T5-large-SQuAD-ChroniclingAmericaQA:} fine-tuned on both SQuAD 1.1 and ChroniclingAmericaQA datasets.
\end{itemize}

\begin{table*}
\caption{Model Performance of ChroniclingAmericaQA Dataset using Corrected OCR Paragraphs and Raw OCR Paragraphs as context. Red colored numbers denote percentage decrease when using Raw OCR paragraphs. The underlined results depict which variation of the model gave the best performance compared to its counterparts, whereas the bold results show the overall best performance on the given measure. }
\label{tab:model_performance}
\resizebox{0.7\linewidth}{!}{
\begin{tabular}{lcccc}
\toprule
\textbf{Model} & \multicolumn{2}{c}{\textbf{Corrected OCR Paragraph}} & \multicolumn{2}{c}{\textbf{Raw OCR Paragraphs}} \\
\cmidrule(lr){2-3} \cmidrule(lr){4-5}
               & \textbf{EM} & \textbf{F1} & \textbf{EM} & \textbf{F1} \\
\midrule
BERT-base                        & 0.12                     & 2.91 & 0.08   \textcolor{red}{(33\%$\downarrow$)}& 2.27  \textcolor{red}{(22\%$\downarrow$)} \\
BERT-base-SQuAD                  & 44.70                    & 57.24 & 27.14 \textcolor{red}{(39\%$\downarrow$)} & 40.33 \textcolor{red}{(29\%$\downarrow$)} \\
BERT-base-ChroniclingAmericaQA   & 63.29                    & 69.43 & 38.94 \textcolor{red}{(38\%$\downarrow$)} & 48.65 \textcolor{red}{(30\%$\downarrow$)}\\
BERT-base-SQuAD-ChroniclingAmericaQA & \underline{63.90}    & \underline{69.92} & \underline{39.71} \textcolor{red}{(37\%$\downarrow$)} & \underline{49.45} \textcolor{red}{(29\%$\downarrow$)}\\
\midrule
RoBERTa-base                     & 0.02                     & 1.48 & 0.02  \textcolor{red}{(0\%$\downarrow$)} & 1.32  \textcolor{red}{(10\%$\downarrow$)} \\
RoBERTa-base-SQuAD               & 47.52                    & 60.55 & 29.83 \textcolor{red}{(38\%$\downarrow$)} & 42.71 \textcolor{red}{(29\%$\downarrow$)}\\
RoBERTa-base-ChroniclingAmericaQA & 63.61                   & 70.72 & 40.40 \textcolor{red}{(36\%$\downarrow$)}& \textbf{\underline{50.12}} \textcolor{red}{(29\%$\downarrow$)}\\
RoBERTa-base-SQuAD-ChroniclingAmericaQA & \underline{63.93} &\textbf{ \underline{71.00}} & \textbf{\underline{40.42}} \textcolor{red}{(36\%$\downarrow$)} & \textbf{\underline{50.12}} \textcolor{red}{(29\%$\downarrow$)} \\
\midrule
T5-large                         & 46.69                    & 59.77 & 26.66 \textcolor{red}{(43\%$\downarrow$)} & 36.90 \textcolor{red}{(38\%$\downarrow$)} \\
T5-large-SQuAD                   & 1.50                     & 9.28  & 0.8   \textcolor{red}{(46\%$\downarrow$)}& 6.07  \textcolor{red}{(34\%$\downarrow$)} \\
T5-large-ChroniclingAmericaQA    & \textbf{\underline{64.28}}           & \underline{69.65} & \underline{34.49} \textcolor{red}{(46\%$\downarrow$)} & \underline{41.54} \textcolor{red}{(40\%$\downarrow$)} \\
T5-large-SQuAD-ChroniclingAmericaQA  & 64.24                & 59.77 & 34.30 \textcolor{red}{(47\%$\downarrow$)} & 41.53 \textcolor{red}{(18\%$\downarrow$)}\\
\bottomrule
\end{tabular}
}
\end{table*}

\begin{figure}[h!]
\includegraphics[width=0.95\columnwidth]{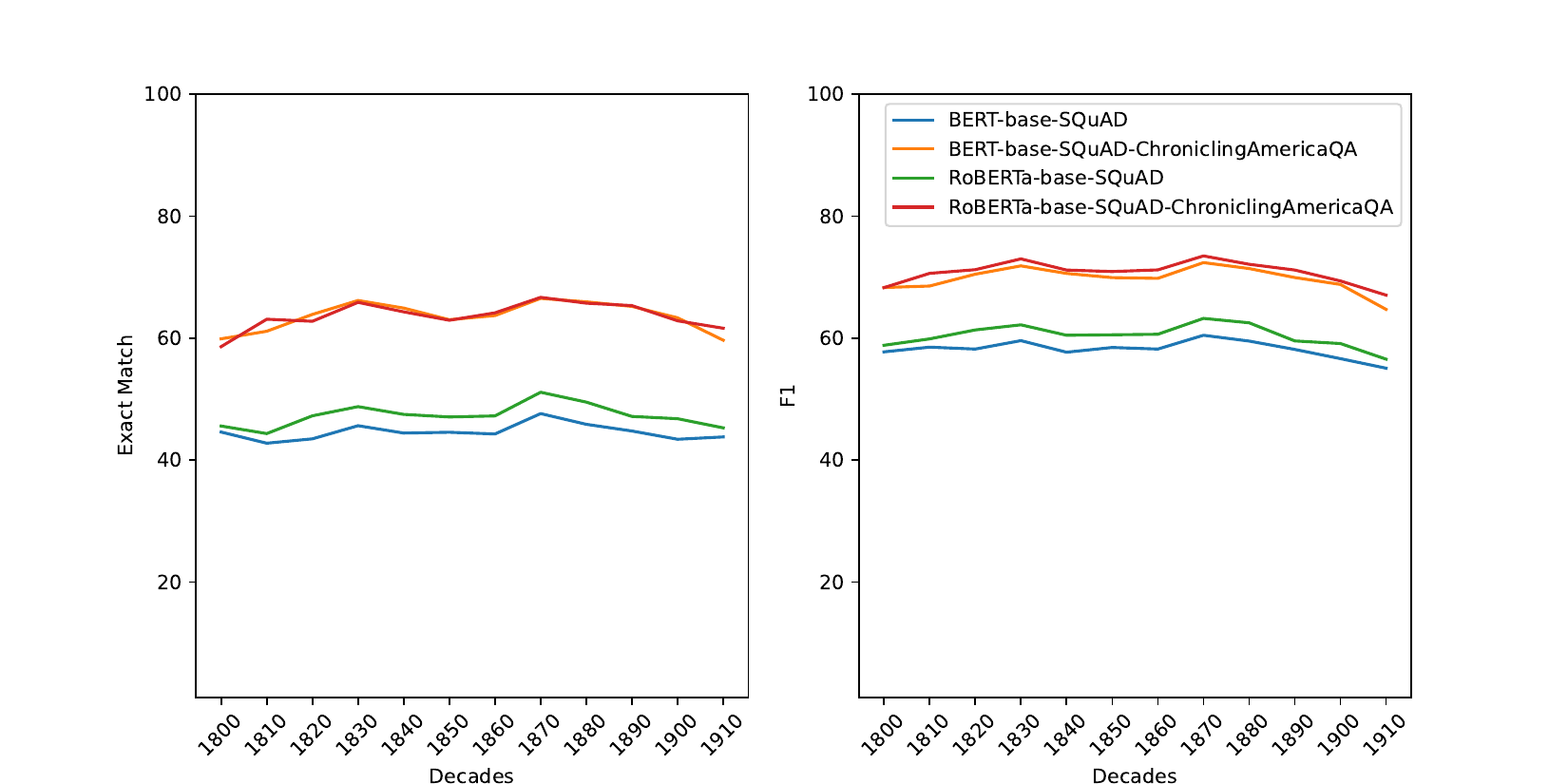}
\caption{Comparison of Model Performance over Time. }
\label{fig:ComparisionModelPerformance}
\end{figure}

Table \ref{tab:model_performance} summarizes the performance of different transformers on the ChroniclingAmericaQA dataset using standard QA metrics: Exact Match (EM) and F1 score.  We measure how well different models can answer questions from both Corrected and Raw OCR paragraphs.
From the results, it is evident that models used without any fine-tuning performed the worst, as these models lack specific knowledge about the question-answering task. Conversely, models fine-tuned on the SQuAD 1.1 dataset show significant improvements, highlighting the benefits of fine-tuning on task-specific datasets. Notably, the highest performance levels are observed from the models trained on the ChroniclingAmericaQA dataset itself, depicting the importance of fine-tuning on specific datasets. However, models trained on both SQuAD and ChroniclingAmericaQA give even better performance, suggesting that combining domain-specific knowledge such as historical content and general task specific knowledge can enhance the performance of models. 

We also compare the results of transformer models on both Raw OCR paragraphs and Corrected OCR paragraphs in Table \ref{tab:model_performance}. From the results, we can see that model performance significantly drops when the Raw OCR paragraph is given as context, suggesting the importance of cleaning the digitized text. We also show the performance of fine-tuned models over time in Figure \ref{fig:ComparisionModelPerformance}. It can be concluded from the figure that the performance improvement due to finetunning on our dataset remains relatively stable over time.

In summary, we can conclude that results shown in Table \ref{tab:model_performance} imply that fine-tuning models on relevant datasets significantly improves the performance. Additionally, we can also conclude that correcting the OCRed text is crucial for getting better model performance. 

\subsection{LLM Performance}
\label{s:LLMperformance}

\begin{table*}[ht!]
\caption{LLMs Performance on ChroniclingAmericaQA using Corrected OCR Paragraphs and Raw OCR Paragraphs as context. Red colored numbers denote percentage decrease when using Raw OCR paragraphs. The bold values represent overall best performance for a given measure on Corrected OCR Paragraphs, whereas the underlined values represent the best performance for the given measure on Raw OCR Paragraphs. }
\resizebox{0.7\linewidth}{!}{
\label{tab:model_performance_LLM}
\begin{tabular}{lclllll}
\toprule
\textbf{Model}  & \textbf{Parameter} & \textbf{EM} & \textbf{F1} & \textbf{Recall} & \textbf{Precision} & \textbf{Contains} \\
\midrule
LLaMA2  (Cleaned OCR Paragraphs) &7B & 0.22 & 12.04 & 57.26 & 7.08 & 44.39 \\
LLaMA2  (Raw OCR Paragraphs) &7B & 0.00 \textcolor{red}{ (100\%$\downarrow$)} & 2.24 \textcolor{red}{ (81\%$\downarrow$)} & 10.45 \textcolor{red}{ (82\%$\downarrow$)} & 1.32 \textcolor{red}{ (81\%$\downarrow$)} & 5.67 \textcolor{red}{ (87\%$\downarrow$)} \\
\midrule
LLaMA2 (Cleaned OCR Paragraphs) &13B& 0.00 & 11.89 & 58.36 & 6.86 & 45.80 \\
LLaMA2 (Raw OCR Paragraphs)&13B & 0.00 \textcolor{red}{ (0\%$\downarrow$)} & 2.16 \textcolor{red}{ (82\%$\downarrow$)} & 10.59 \textcolor{red}{ (81\%$\downarrow$)} & 1.27 \textcolor{red}{ (81\%$\downarrow$)} & 5.68 \textcolor{red}{ (87\%$\downarrow$)} \\
\midrule
LLaMA2 (Cleaned OCR Paragraphs) &70B& \textbf{5.30} & \textbf{19.52} & 61.38 & \textbf{14.17} & 48.37 \\
LLaMA2 (Raw OCR Paragraphs) &70B& \underline{0.39} \textcolor{red}{ (93\%$\downarrow$)} & \underline{3.02} \textcolor{red}{ (84\%$\downarrow$)} & \underline{11.19} \textcolor{red}{ (82\%$\downarrow$)} & \underline{2.00} \textcolor{red}{ (86\%$\downarrow$)} & \underline{6.35} \textcolor{red}{ (87\%$\downarrow$)} \\
\midrule
\midrule
Mixtral (Cleaned OCR Paragraphs) &8x7B& 1.13 & 12.88 & \textbf{62.69} & 8.15 &\textbf{ 50.00} \\
Mixtral (Raw OCR Paragraphs) &8x7B & 0.00 \textcolor{red}{ (100\%$\downarrow$)} & 0.94 \textcolor{red}{ (92\%$\downarrow$)} & 10.06 \textcolor{red}{ (84\%$\downarrow$)} & 0.51 \textcolor{red}{ (94\%$\downarrow$)} & 5.63 \textcolor{red}{ (89\%$\downarrow$)} \\
\midrule
Mistral (Cleaned OCR Paragraphs) &7B & 1.17 & 16.30 & 56.35 & 10.42 & 44.16 \\
Mistral (Raw OCR Paragraphs) &7B & 0.09 \textcolor{red}{ (92\%$\downarrow$)} & 2.08 \textcolor{red}{ (87\%$\downarrow$)} & 8.19 \textcolor{red}{ (85\%$\downarrow$)} & 1.30 \textcolor{red}{ (85\%$\downarrow$)} & 4.33 \textcolor{red}{ (90\%$\downarrow$)} \\
\bottomrule
\end{tabular}%
}
\end{table*}

Similarly, in addition to transformer models, we evaluate our dataset with Large Language Models (LLMs), including the LLaMA2 \cite{touvron2023llama} family, Mixtral 8x7B \cite{jiang2024mixtral} and Mistral 7B \cite{jiang2023mistral}. Evaluating the performance of LLMs \cite{guo2023evaluating} within the context of question answering poses unique challenges. Traditionally, QA systems are assessed using Exact Match (EM) and F1 scores. However, these metrics are not fully appropriate for generative models such as LLMs, which often produce verbose responses. This verbosity can result in an EM value of zero due to the presence of extra tokens in the answer. Similarly, the F1 score, which combines both recall and precision, may not provide an accurate measure of performance under these conditions. To overcome the challenge of evaluating LLMs, we introduce model-agnostic metrics like Token Recall and Answer String Containment\footnote{\url{https://huggingface.co/spaces/evaluate-metric/squad}}, which are more suitable for assessing verbose responses. Token Recall evaluates how effectively the model's response covers the information found in the ground truth, while Answer String Containment assesses the degree to which the ground truth is included within the model's response. By using these metrics, we aim to offer a more equitable evaluation of an LLM's ability to generate responses to questions.

Table \ref{tab:model_performance_LLM} presents the performance of different LLMs on the ChroniclingAmericaQA Dataset using both raw OCR and corrected paragraphs. When evaluating the results of different LLMs, we can note that LLaMA2 70B gives significantly higher EM and F1 scores compared to its counterparts and Mistral models. This also suggests that models with large sizes have a greater ability to understand the nuances of historical text, leading to more accurate and precise answers. 
Meanwhile, the Mixtral 8x7b model performs well on recall and contains metrics, depicting its proficiency in extracting relevant information from the context. 

Furthermore, when examining the results of different LLMs considering the raw OCR paragraphs and corrected paragraphs, we notice that performance across all models decreased drastically when utilizing the raw OCR paragraphs as context. Models such as LLaMA2 7B and Mixtral Bx7B gave zero results on EM metric, whereas LLaMA2 70B still showed some resistance on raw OCR paragraphs, however, with a significant drop of 93\%. Moreover, when analyzing the recall and precision metrics results, one can notice that models showed better performance on recall, indicating their ability to capture relevant information from the context.

Summarizing the results of Table \ref{tab:model_performance_LLM}, we can conclude that the performance of LLaMA2 70B is better compared to other models when handling both raw OCR text and corrected text. On the other hand, Mixtral 8x7B gives good results for recall and contains metrics on corrected OCR paragraphs. However, LLaMa2 70B has a high recall and contains values on the raw OCR paragraphs.

Overall, we can conclude that the results of both Tables \ref{tab:model_performance} and \ref{tab:model_performance_LLM} emphasize that fine-tuning the model on the relevant datasets plays a major role in increasing the model performance. Additionally, it highlights the necessity of post-processing OCR text for improving the performance of models on historical documents, as the results of both transformer models and LLMs degrade drastically for raw OCR text.

\subsection{Human Evaluation}
\label{s:Human_evaluation}

To assess the quality of the ChroniclingAmericaQA dataset, we also conducted a manual evaluation study. We randomly selected 360 question-answer pairs for evaluation, sampling 30 pairs from each decade represented in the dataset. Six graduate students were asked to rate these pairs on a scale of 1 to 5, with 1 being "very bad" and 5 being "very good."  The evaluators were instructed to assess the pairs based on four criteria: the readability of the question, ensuring it is grammatically correct and flows well; the readability of the answer, verifying its grammatical correctness and fluency; relevance, to ascertain if the generated question pertains to the provided passage; and non-ambiguity, to ensure the question is straightforward and clear. Each question was annotated by one of the annotators.

Each criterion's results were averaged and are presented in Table~\ref{tab:humaneval_results}. The evaluations gave high scores across all metrics, indicating that our dataset is of good quality, particularly regarding the relevance and readability of both questions and answers. The score for non-ambiguity was also notably high, demonstrating that most questions are direct and unambiguous.

\begin{table}[]
\caption{Human evaluation results.}
\resizebox{.8\columnwidth}{!}{%
\label{tab:humaneval_results}
\begin{tabular}{|c|c|c|c|}
\hline
\textbf{\makecell{Readability of \\ Question}} & \textbf{\makecell{Readability \\ of Answer}} & \textbf{Relevance} & \textbf{\makecell{Non-Ambiguity}}  \\
\hline
4.24                    & 4.29                   & 4.39      & 4.18           \\   
\hline
\end{tabular}%
}
\end{table}

\section{Use cases}
\label{s:DatasetUseandLimitaion}
Our dataset can be used in several ways. First, it provides a new benchmark for training and evaluating QA and MRC models on historical texts, enabling models to handle the complexities of historical documents, including OCR inaccuracies and language evolution. The evaluation can be done for arbitrary, different spans of time within the time frame of the dataset (1800-1920). It can be also conditioned on different regions of USA or particular states. 

Second, since our dataset contains scanned images of news papers as well as the corresponding raw and cleaned OCR texts, it is possible to benchmark models on each of these. Some memory institutions still hold only scans of historical materials, while others, such as in the case of the Chronicling America, may contain quite noisy OCRed text. Our dataset allows then benchmarking models on these different types of inputs, offering realistic scenarios in which the models may need to be required to operate.

Third, our dataset can be utilized in a similar manner to datasets such as SQuAD \cite{rajpurkar-etal-2016-squad}, ArchivalQA \cite{archivalqa-wang}, and StreamingQA \cite{liska2022streamingqa} for contributing to the ongoing efforts in enhancing the adaptability and knowledge retention of large language models \cite{hu-etal-2023-meta, tack2024online}.

%


Finally, ChroniclingAmericaQA could potentially serve as an educational resource or as a tool to increase the public's engagement with historical documents; for example, it could be used by teachers and educators to assess students' reading comprehension on rare, historical materials. Note, that our dataset generation framework can be applied on other historical news collections.

\section{Conclusion}\label{s:Conclusion}

In this work, we introduce ChroniclingAmetricaQA, a large-scale question-answering dataset comprising 487k question-answer pairs over a collection of historical American newspapers with the objective of facilitating the development of QA and MRC systems over historical texts. Our dataset is unique as it is specifically based on historical newspapers covering the longest time period of 120 years (from 1800 to 1920) among available datasets based on news documents. We highlight the difficulties of utilizing historical newspapers for QA tasks, such as the poor quality of OCR text and the need for post-processing to enhance text quality. We employed language models such as GPT 3.5 Turbo for OCR text correction and the T5-base model for question generation to mitigate these challenges. We also carried out a comprehensive evaluation of our dataset on various transformer models and LLMs, demonstrating the effectiveness of our dataset and providing reference baselines. Models trained on ChroniclingAmericaQA were found to consistently surpass the performance of those fine-tuned on other datasets.

\textbf{Ethical considerations} Lastly, we would like to mention ethical considerations of our work. Since the dataset has been created from temporally distant data, there is possibility that some questions may be offensive or biased in relation to some ethnic groups or other segments of society. We made sure that the possibility of such questions occurring in our dataset is low by performing post-processing search using a set of prepared keywords coupled with careful manual analysis. This kind of issue is not specific to our dataset but is a problem typical in many heritage materials \cite{borenstein-etal-2023-measuring}, including also ones in the Chronicling America repository, and should be continuously studied and mitigated in the future.

\bibliographystyle{ACM-Reference-Format}
\bibliography{sample-base}

\appendix

\end{document}